%% file: main.tex
\pdfoutput=1

\documentclass[conference]{ieeeconf}
\IEEEoverridecommandlockouts
\usepackage[bookmarks=false]{hyperref}

\usepackage{url}

\usepackage{graphicx}
\usepackage{xcolor}
\usepackage{lipsum}
\usepackage{url}

\usepackage[skip=2pt,font=small]{caption}
\usepackage{subcaption}

\usepackage{graphicx}
\usepackage{multirow}
\usepackage{threeparttable}
\usepackage{placeins}
\usepackage{amsmath}
\usepackage{algorithm}
\usepackage[noend]{algpseudocode}
\usepackage{booktabs}

\usepackage{blindtext}
\usepackage{footnote}
\usepackage[backend=bibtex,
            hyperref=false,
            url=false,
            isbn=false,
            doi=false,
            backref=false,
            style=ieee,
            citestyle=numeric-comp,
            sorting=nyt,%none
            block=none]{biblatex}

\usepackage[utf8]{inputenc}

\definecolor{cadmiumgreen}{rgb}{0.0, 0.42, 0.24}

\usepackage{marginnote}
\usepackage{soul} %text highlighting
\newcommand{\ignore}[1]{}

\newcommand{\dataset}{\mathcal{D}}

\newcommand{\latent}{\mathbf{z}}

\newcommand{\piparams}{\mathbf{\theta}_\pi}
\newcommand{\repparams}{\mathbf{\theta}_s}

\newcommand{\markov}{\mathcal{M}}

\newcommand{\states}{\mathcal{S}}
\newcommand{\actions}{\mathcal{A}}

\newcommand{\taction}{\mathbf{a}_t}
\newcommand{\stateinitial}{\rho_0}
\newcommand{\rewardfn}{r}
\newcommand{\transition}{\mathcal{T}}
\newcommand{\timemax}{T}
\newcommand{\policy}{\pi}

\newcommand{\cartdelta}{\Delta \mathbf{x}}

\newcommand{\varpoe}{\mathbf{\sigma}^2}
\newcommand{\varpoee}{\mathbf{\sigma}^{2}}
\newcommand{\mupoe}{\mathbf{\mu}}
\begin{document}

% brainstorming title:
%\title{\LARGE {\bf Detect and Correct}: Out-of-Distribution Detection for Crossmodal Compensation of Corrupted Sensor Data}

% \title{\LARGE {\bf
% Detect and Correct}: Leveraging Out-of-Distribution Detection for Crossmodal Compensation of Corrupted Sensors}

\title{\LARGE {\bf
Detect, Reject, Correct}: Crossmodal Compensation of Corrupted Sensors}

\author{Michelle A. Lee$^1$, Matthew Tan$^{1}$, Yuke Zhu$^{2}$, Jeannette Bohg$^{1}$
\thanks{$^{1}$Stanford University, %
$^{2}$The University of Texas at Austin
% {\footnotesize [mishlee, , bohg]@stanford.edu}
}

\thanks{This work has been supported by NSF CNS-1955523 and JD.com American Technologies Corporation (“JD”) under the SAIL-JD AI Research Initiative. This article solely reflects the opinions and conclusions of its authors and not JD or any entity associated with JD.com. We are grateful to Mike Salvato and Brent Yi for providing invaluable feedback, and to Peter Zachares for his help in developing the simulation environment.}
}

\maketitle
\begin{abstract}
%\begin{quote}

    Using sensor data from multiple modalities presents an opportunity to encode redundant and complementary features that can be useful when one modality is corrupted or noisy. Humans do this everyday, relying on touch and proprioceptive feedback in visually-challenging environments. However, robots might not always know when their sensors are corrupted, as even broken sensors can return valid values. In this work, we introduce the Crossmodal Compensation Model (CCM), which can detect corrupted sensor modalities and compensate for them. CMM is a representation model learned with self-supervision that leverages unimodal reconstruction loss for corruption detection. CCM then discards the corrupted modality and compensates for it with information from the remaining sensors. We show that CCM learns rich state representations that can be used for contact-rich manipulation policies, even when input modalities are corrupted in ways not seen during training time.

    % Deep learning provides a powerful tool for robot perception and controls from multimodal raw sensory inputs. However, in real-world robotic systems, sensors can become occluded, be knocked out of their calibrated positions, or even break during deployment. This requires a system that can detect 

    % Crossmodal compensation is an incredibly challenging problem due to the non-intuitive, complex relationships between different modalities. 
    % Using multiple sources of information presents an opportunity to encode redundant, complementary features across modalities that can be useful when one modality is lost. Humans do this everyday.  Crossmodal compensation is an incredibly challenging problem due to the non-intuitive, complex relationships between different modalities. 
    % Using multiple sources of information presents an opportunity to encode redundant, complementary features across modalities that can be useful when one modality is lost. Humans do this everyday. 

%\end{quote}
\end{abstract}

\input{1-intro.tex}

\input{2-relwork}

\input{3-approach}
\input{4-expt}

\input{5-disc}

\printbibliography %for biblatex

\end{document}

%% file: 1-intro.tex
\section{Introduction}
% I like it as well ... But if we remove it, nothing is really missing from the story
%Here is an experiment that you can try at home: take a water bottle and unscrew the cap. Now close your eyes and try to close the water bottle. Most humans, even when they lose their visual senses, can rely on proprioceptive and tactile sensing when performing manipulation tasks. To study the inverse relationship (in an experiment best not done at home), Johansson et  al.~\cite{johansson2009coding} anesthetized the fingertips of human subjects which impacted their ability to light a match. The experiment~\cite{youtube} shows that while the human subject at first struggled to manipulate the match without haptic feedback, they were still able to light it within about 20 seconds of trial and error. 

Here is an experiment that you can try at home: take a water bottle and unscrew the cap. Now close your eyes and try to close the water bottle. Most humans, even without visual senses, can rely on proprioceptive and tactile sensing when performing manipulation tasks. To study the inverse relationship (in an experiment best not done at home), Johansson et  al.~\cite{johansson2009coding} anesthetized the fingertips of human subjects which impacted their ability to light a match. The experiment video~\cite{youtube} shows that while the human subject at first struggled to manipulate the match without haptic feedback, they were still able to light it within about 20 seconds of trial and error.  Neuroscience research provides more evidence that humans can take information from one sensor modality to compensate for missing information of another in a {\em crossmodal\/} manner. %Crossmodal refers to the interaction between two sensor modalities. 
This has been shown for visual and tactile feedback in tasks, such as object size prediction~\cite{ernst2002humans} and object manipulation~\cite{jenmalm1997visual}, as well as visual and auditory information~\cite{lomber2010cross, baart2017cross, alais2004ventriloquist}.

% I don't think it hurts to remove this. You will tlak about this later anyway when you formulate the problem
%Many works have shown that machine learning models fail catastrophically when performing inference on noisy or {\em out-of-distribution\/} (OOD) data not seen during training time~\cite{sugiyama2012machine, amodei2016concrete, goodfellow2014explaining}. As learning-based methods are deployed more and more often on real robotic system, it is important to avoid potentially unsafe behaviour due to OOD data~\cite{evtimov2017robust}.

We aim to endow a robot with the same capability of crossmodal compensation. This is an important for avoiding potentially dangerous outcomes when deploying a robot to the real world. There are many cases when a sensor can break, produce erroneous data, become occluded, or change with lighting conditions. In this work, we focus on the case
%on a specific subset of out-of-distribution data 
in which one of our sensors experiences failure modes or noise unseen during train time. We want our robot to accomplish tasks robustly, even in the face of the corrupted sensor data. 

There have been several recent works that can perform inference or complete downstream task  in the presence of missing modalities~\cite{wu2018multimodal, tan2019factorized, tsai2018learning, liu2017learning,li2019connecting, takahashi2019deep}. However, they need to know what modality is missing at inference time. In this work, we are interested in crossmodal compensation at inference time, when we lack knowledge on which sensor modality may be corrupted. 

% Examples of OOD modalities can be a camera trying to capture RGB information when someone turns off the light, a depth camera that is knocked over so the inputs are now skewed, or a force sensor with broken strain gauges. 
Other works, particularly in autonomous driving, actively detect when sensor inputs are {\em out-of-distribution\/} (OOD) with respect to the training data ~\cite{kahn2017uncertainty,richter2017safe} and compensate for it \cite{mcallister2019robustness, filos2020can}. However, these works focus on avoiding collision and require querying expert feedback, which limit the generalization of their methods to manipulation tasks. 

\begin{figure}[t]
\centering
\includegraphics[width=0.5\textwidth,trim={0.0cm 4.0cm 0.0cm 0cm},clip]{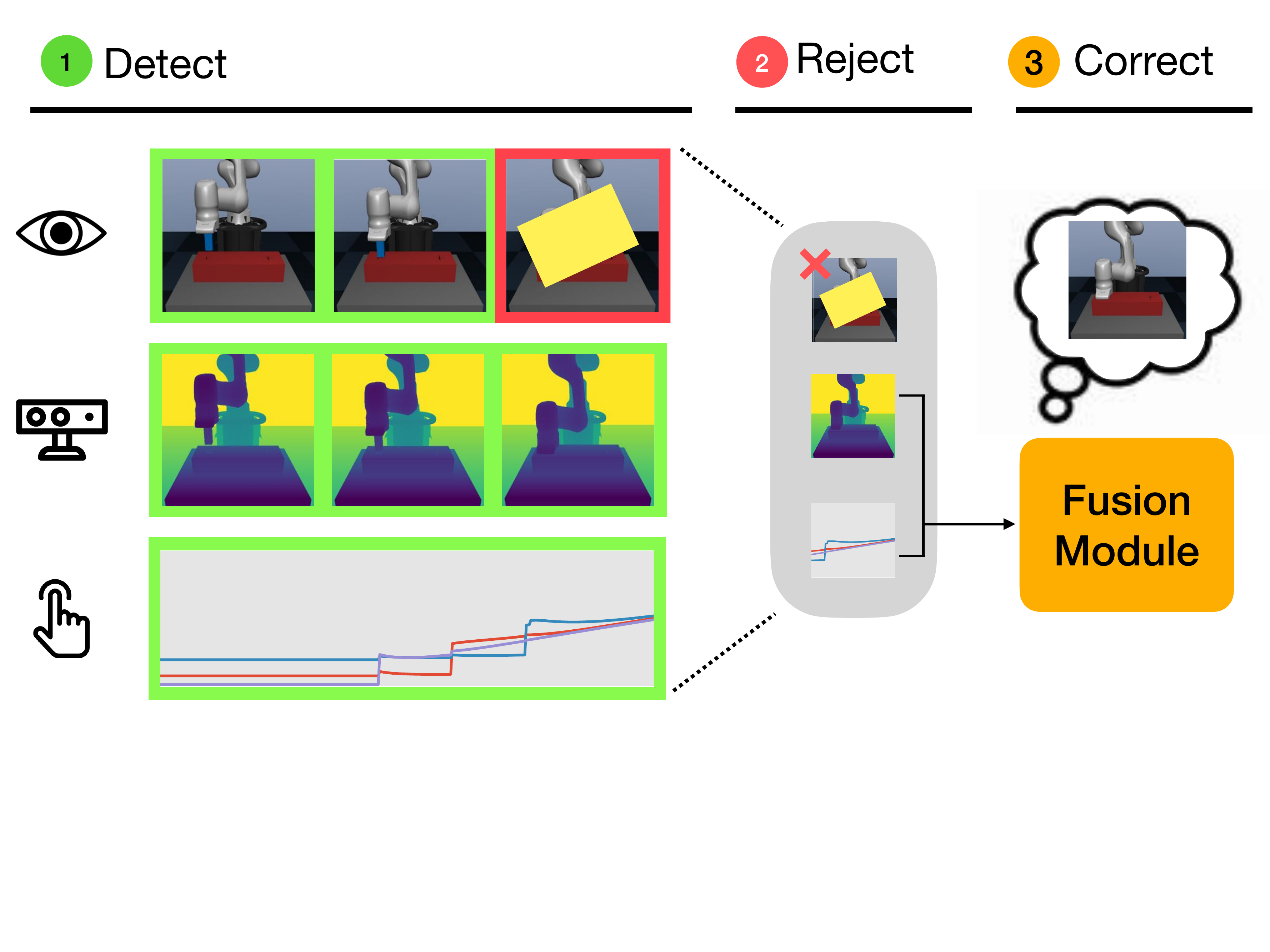}
\caption{
We propose Crossmodal Compensation Model that can (1) \textbf{detect} when a sensor input is corrupted, which in this case is the image data, (2) \textbf{reject} the corrupted image as input to our policy, and (3) \textbf{correct} for the rejected image by compensating the missing image information with the remaining force and depth information in the fusion module.
}
\label{fig:pull}
\end{figure}

Several works on robot manipulation have shown that Bayesian filtering approaches to sensor fusion can perform state estimation even when some sensor readings are corrupted~\cite{martin2017cross, martin2017building, yu2018realtime, lambert2019joint}. These methods require the user to explicitly define the estimated state, as well as the analytical forward and measurement models that may be hard to specify or intractable to compute online. While differentiable filters such as~\cite{lee2020multimodal} can learn how to fuse multiple modalities, they were not demonstrated to recover from corrupted sensor inputs unseen during train time. 

To detect corrupted sensor readings {\em and\/} compensate for them, we introduce the {\em Crossmodal Compensation Model\/} (CCM), a novel latent variable representation model that can be used to generate state feedback for a learned policy that compensates for corrupted sensor inputs. CCM performs crossmodal compensation for corrupted sensor inputs in three steps: CCM (1) \textbf{detects} which modality is corrupted through out-of-distribution detection, (2) \textbf{rejects} and discards the corrupted sensor reading, and (3) \textbf{corrects} for the discarded modality with information from the remaining modalities. By learning to reconstruct each input modality, CCM compares unimodal reconstruction results with the sensor inputs to identify the corrupted sensor. CCM jointly learns this reconstruction objective with self-supervised objectives introduced in~\cite{lee2019making} that were shown to be crucial for learning a rich representation that lends itself to control. CCM learns to perform crossmodal compensation by minimizing the distance between a representation generated with dropped modalities and the representation generated with full modalities. CCM is robust to corrupted sensor readings, and, by correcting for them, generates rich state representations that can be used for contact-rich manipulation policies.

%% file: 2-relwork.tex
\section{Related Work}

\noindent \textbf{Multimodal Representation Learning.} The complementary nature of heterogeneous sensor modalities holds the promise of providing more informative feedback for solving perception and manipulation tasks than uni-modal approaches. Several works have combined visual and tactile information for object tracking~\cite{martin2017cross, lambert2019joint, yu2018realtime}, shape completion~\cite{wang20183d}, grasp assessment \cite{Calandra:2018,gao2016deep,YaseminRenaud,SinapovSS14}, and scene understanding~\cite{bohg2010strategies, bohg2011multi}. Many of these multimodal approaches are trained through a classification objective for inference tasks~\cite{Calandra:2018,gao2016deep,YaseminRenaud,yang2017deep}. In prior works \cite{lee2019making, lee2019icra}, we proposed a self-supervised approach to learn multimodal representations that were used as policy inputs, but do not explicitly use the multimodal information to compensate for corrupted sensor readings. In this work, we leverage similar architectures and self-supervised objectives from~\cite{lee2019icra} to train our representations. 
% Here, we build upon that prior work as we are interested in learning representations useful for control and policy learning and leverage the same architectures and forward dynamics objective to train our representations. 

\begin{figure}[t!]
\centering
\includegraphics[width=\linewidth,trim={0cm 5cm 9.8cm 0cm}, clip]{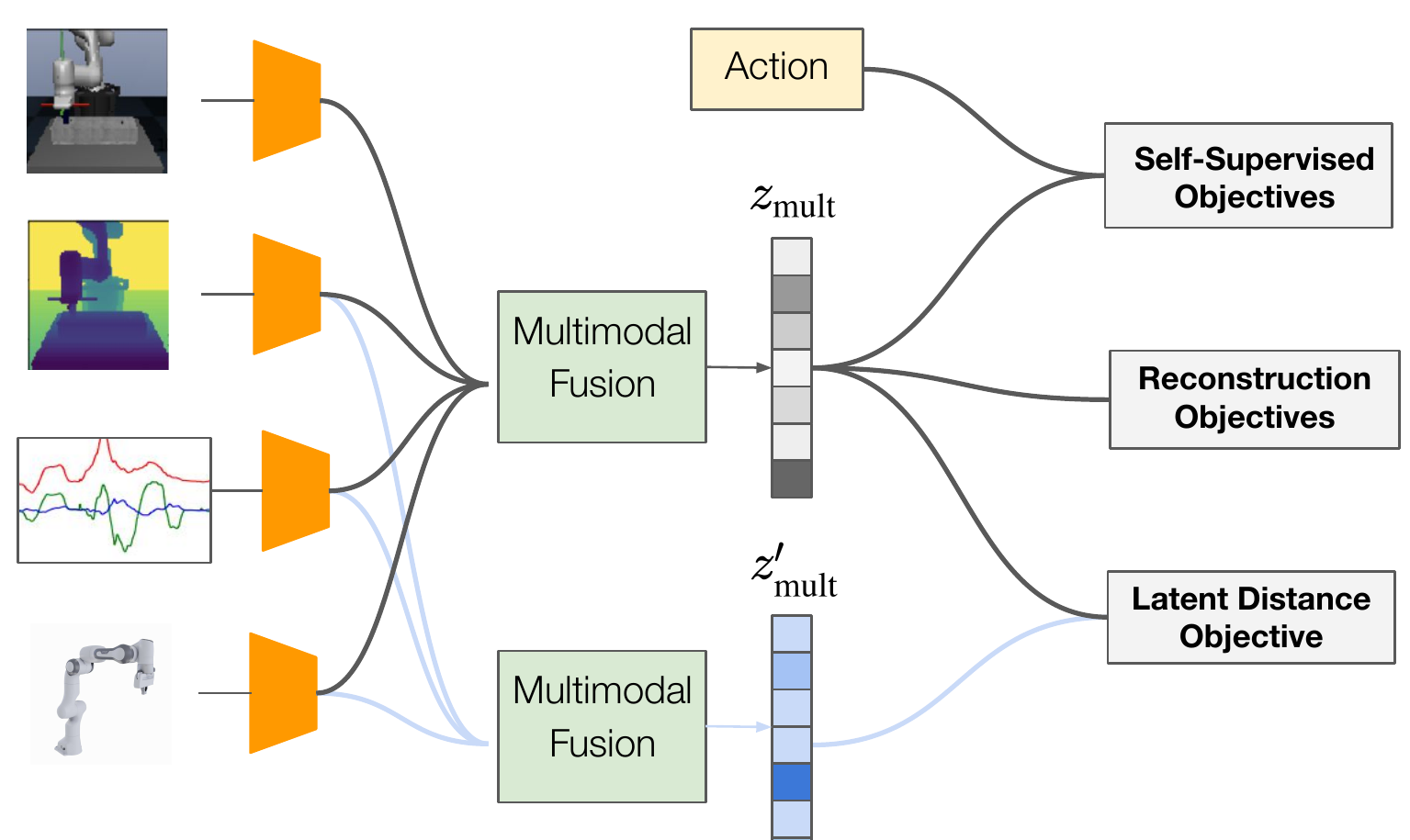}
\caption{
CCM is a multimodal latent representation model that takes RGB image, depth, force, and robot proprioception data as inputs, and jointly trains to reconstruct the input modalities as well as to predict self-supervised objectives used in~\cite{lee2019making}. Additionally, we learn a latent distance objective: during training, we drop a random input modality and learn to reduce the L2 distance between our compensated latent representation $z'_{mult}$ and the full modality representation $z_{mult}$.
}
\label{fig:graphical_model}
\end{figure}

\noindent \textbf{Missing Modality Representations.} While many works have shown that using multimodal data sources can improve the robustness and accuracy of an inference task~\cite{wu2018multimodal, tsai2018learning, lee2019making, baltruvsaitis2018multimodal, zambelli2020multimodal, pham2019found}, the multimodal data source can sometimes be noisy or incomplete. Motivated by this, previous works have tackled the problem of noisy or missing modalities by learning explicit crossmodal models or robust joint representations.

One approach to handle missing modalities is to explicitly predict how to transfer from one modality to another. \cite{li2019connecting, takahashi2019deep} use deep generative models to predict between raw vision and touch. Other works, such as \cite{tsai2018learning, wu2018multimodal, liu2017learning, tan2019factorized}, drop modalities during train time to reconstruct the missing modality, perform inference, or both.  

To account for the missing modalities during training, Tsai et al.~\cite{tsai2018learning} learns a new model for each missing modality. On the other hand, Wu et al.~\cite{wu2018multimodal} and Tan et al.~\cite{tan2019factorized} use a variational product-of-experts (PoE) approach to recover the joint representation when inputs are missing, eliminating the need to parameterize new models. Our work uses the PoE approach similar to \cite{wu2018multimodal}. All recent work on compensating missing modalities assume to know which modality is missing, which is not always the case. In contrast, our approach both detects and compensates for missing modalities. 

\noindent \textbf{Out of Distribution Detection.} Detecting OOD inputs in control tasks can prevent catastrophic failure in robots. While many works detect OOD data with deep neural network-based architectures~\cite{filos2020can, richter2017safe, mcallister2019robustness, kahn2017uncertainty}, none of these work look at using crossmodal compensation for OOD data. In this work, we follow~\cite{richter2017safe} and train a variational autoencoder, using the reconstruction error at test time to detect when a single modality is OOD. There is a vast amount of literature relating to OOD detection, and we refer the reader to this survey paper~\cite{chalapathy2019deep} for more information.

% We extend \cite{richter2017safe} by presenting a simple secondary threshold to determine which modality is most OOD when multiple modalities are detected as OOD. 

%% file: 3-approach.tex
\section{Problem Statement}
Our goal is to learn to perform a manipulation tasks even with corrupted sensor readings. Our algorithm learns the manipulation task with multiple modalities as input, and uses data from the un-corrupted sensor modalities to compensate for the corrupted ones. 
% In other words, our algorithm is able to leverage the crossmodal information from uncorrupted modalities to be robust to corrupted sensor input. 
% We first learn a representation of the multisensory data that can detect when sensor inputs are corrupted and \jean{too fuzzy. What do you mean exactly by compensate? Mention that you detect the corrupted input, then you remove it. And then the resulting latent z still should be close to the one when the sensor input wasn't corrupted.} compensate for the corrupted sensor input. We use the learned latent representation, which is \jean{robust here is fuzzy. Describe what you mean! You mean (I think) that the latent z is the same w/o modality and w/ uncorrupted modality} robust to corrupted sensor readings, as input to our policy learned with reinforcement learning. \mish{Isn't the detect & remove our method? The problem formulation is: you have corrupted input and your policy needs to do something about it. Our method is to throw out the corrupted input}
We model the manipulation task as a finite-horizon, discounted Markov Decision Process (MDP) $\markov$, with a state space $\states$, an action space $\actions$, state transition dynamics $\transition : \states \times \actions \to \states$, an initial state distribution $\stateinitial$, a reward function $r : \states \times \actions \to R$, horizon $\timemax$, and discount factor $\gamma \in (0, 1]$. To determine an optimal policy $\policy$, we want to maximize the expected discounted reward $E_\policy \left[\sum^{\timemax-1}_{t=0} \gamma^t \rewardfn(s_t, \taction) \right]$.

We represent the policy by a neural network with parameters $\piparams$ that are learned as described in Sec.~\ref{sec:policy-control}. $\actions$ is defined over continuously-valued 3D displacements $\cartdelta$ in Cartesian space. $\states$ is defined by the low-dimensional latent representation $z_{mult} = f(o_1, o_2, ...o_n)$ inferred from multimodal sensory inputs $o_i$ through an encoder $f$. This encoder is a neural network parameterized by  $\psi_s$ 

In this work, we are interested in cases when the robot is receiving corrupted sensor readings during policy rollout at test time. We describe how our model compensates for corrupted sensor readings below.

\section{Method Overview}
Our proposed model, CCM, attempts to detect and compensate for corrupted sensor readings at the representation level. CCM is a multimodal latent variable model that encodes heterogeneous inputs into a multimodal representation $z_{mult}$ using a variational PoE approach \cite{wu2018multimodal}. We jointly train modality reconstruction, self-supervised objectives, and a latent distance objective to learn useful representations.

To compensate for the corrupted sensor reading $o'_i$, the representation model first detects which sensor is corrupted (see Sec.~\ref{sec:ood-detection}). Then, the representation model removes the corrupted input and performs crossmodal compensation, i.e. inferring a compensated latent representation $ z'_t$ that is close to $z_{mult}$ (fully modality latent representation): $f(o_1, o_2, ... o_i', ...o_n) = \hat z'_{mult}\approx z_{mult}$(see Sec.~\ref{sec:crossmodal-compensation}). 

% For detection, we compare the input data with its reconstruction as generated by our representation model. High reconstruction errors indicate that the corresponding input is out of the training data distribution. We describe OOD detection in Section \ref{sec:ood-detection}. After detecting a corrupted input, we drop the corrupted modality and treat the crossmodal compensation problem as a missing modality problem. We use our variational encoder to recalculate the joint representation with the remaining modalities. This is described in detail in Section \ref{sec:crossmodal-compensation}.

% \begin{figure}[t!]
% \centering
% \includegraphics[width=\linewidth,clip]{images/encoder_decoder.png}
% \caption{Neural network architecture for multimodal representation learning. The network takes data from four different sensors as input: RGB images, depth map, F/T readings, and end-effector position, and velocity. These inputs are encoded using modality specific encoders and fused using a variational bayesian method. To learn useful features, the representation is trained with action conditional forward dynamics objectives and modality reconstruction objectives.}
% \label{fig:encoder_decoder}
% \end{figure}

\subsection{Multimodal Representation Model}
\label{sec:representation-learning}
CCM is a multimodal latent variable model trained with self-supervision. Our model encodes 4 types of data: RGB images($o_{RGB}$), depth images ($o_{depth}$) from a fixed RGB-D camera, haptic feedback from a wrist-mounted force-torque (F/T) sensor ($o_{force}$), and end-effector position, and linear velocity ($o_{prop}$). Information from each modality is then fused into a single multimodal latent representation $z_{mult}$.

We use the same modality specific encoders as our prior work~\cite{lee2019making} to capture domain-specific features, except for our force encoder, as we found that using a convolution-based architecture helps with force reconstruction. We take the last 32 readings from a F/T sensor as a one-channel ($1\times32\times6$) input into a 5-layer two-dimensional CNN. We add a single fully-connected layer to the end of each modality encoder to map into a $2\times d$-dimensional variational parameter vector with $d=128$ as in~\cite{lee2019making}.

We assume that each modality is conditionally independent given the fused multimodal latent variable representation $z_{mult}$. Each modality encoder maps to a multivariate isotropic Gaussian parametrized by $z_m = \{\mu_{m}, \sigma_{m}\}$ which is then fused using a PoE approach~\cite{wu2018multimodal}. The resulting multivariate Gaussian distribution of the multimodal latent space will have mean $\varpoe_{j}=(\sum_{i=1}^{n+1} \varpoee_{ij})^{-1}$ and variance $\mupoe_{j}=(\sum_{i=1}^{n+1} \mupoe_{ij}\varpoee_{ij})(\sum_{i=1}^{n+1} \varpoee_{ij})^{-1}$, where $n$ is the number of modalities, $\mu_j$  and $\sigma_j^2$ are the variational parameters of the $j$-th dimension. We also add an isotropic multivariate Gaussian prior as an additional expert.

Following \cite{lee2019making}, we train our model using a variational objective by minimizing the Evidence of Lower BOund (ELBO) over our dataset $D = \{\{o_i, y_i, a_i\} | i = 1...n\}$ with observations, $o$, labels for self-supervised objectives, $y$, and actions, $a$: $
  \mathcal{L}_i(\repparams, \phi_s) =   E_{q_{\phi_s}(\latent | \dataset_i)}[\log p_{\theta_s}(\dataset_i | \latent)]
   - \mathrm{KL}[q_{\phi_s}(\latent | \dataset_i)||p(\latent)] 
$. We model the approximate posterior $q_{\phi_s}(z | o_i)$ as a neural network encoder parameterized by $\phi_s$. We model the likelihood $p_{\theta_s}(o_i, y_i, a_i | z)$ with a decoder neural network, parameterized by $\theta_s$. 

\noindent\textbf{Decoder Architectures} In this work, our model jointly optimizes self-supervised objectives and reconstruction objectives. For the self-supervised objectives, our model learns to predict action-conditional optical flow, next-step end-effector pose, whether the end-effector will be in contact at the next time step, and whether the input modalities are paired. The decoder architectures for these four self-supervised objectives are described in~\cite{lee2019making}. We also learn to reconstruct our input modalities, commonly used for representation learning~\cite{lesort2018state}. Models that learn to reconstruct their inputs can also be used for out-of-distribution detection~\cite{chalapathy2019deep}, which we use for detecting when an input modality is corrupted during test time. 

The force decoder uses a 4-layer deconvolutional decoder for reconstruction. We found it difficult to reconstruct the small and often noisy torques from the force/torque sensor, so our force decoder only reconstructs 3-dimensions forces. 

For reconstructing the RGB and depth image from the the multimodal vector $z_{mult}$, we use a 5-layer deconvolutional decoder for each modality. To encourage our network to learn reconstruction of the robot interacting with the environment (instead of the static background), we take the boolean sum of the robot in image-space throughout the entire dataset to create a mask. At the end of each of our 5-layer depth and image decoders, one deconvolution layer reconstructs the entire image or depth data and another deconvolution layer reconstructs the image or depth data in the masked region. We find that learning to reconstruct both the masked and complete image/depth helps with learning speed and stability. 

% \textbf{Self-Supervised Learning Objectives Architecture}
% The next action which we define as the end-effector motion, is encoded by a 2-layer MLP. The output of the action encoder is concatenated with the multimodal representation and processed by an additional 2-layer MLP, which is used as the input to the action-conditional decoders: flow predictor, contact predictor, pairing predictor, and end-effector position predictor. 
% The flow predictor uses a 4-layer deconvolutional decoder network with upsampling to process the action-conditional feature vector. Following~\cite{flownet1}, we use 4 skip connections. At the end of these 4 layers, one deconvolution layer predicts the unmasked optical flow of the scene and another deconvolution layer predicts the optical flow mask. These two estimates are multiplied element-wise to predict the optical flow of the robot. The predicted optical flow is a $32\times32\times2$ image which is then upsampled to the size of the ground-truth optical flow $128\times128\times2$. The contact predictor and pairing predictors are both 1-layer MLPs that perform binary classification. The end-effector prediction network is a 4-layer MLP.

% \textbf{Reconstruction Decoder Architecture}

% Simply optimizing the multimodal objective over the entire training set, however, has the unfortunate consequence of never training on dropped modalities given a complete dataset with no missing modalities.
\noindent\textbf{Missing Modality Training Objective} CCM learns to perform modality compensation by dropping one of the 3 modalities $\{o_{RGB}, o_{Force}, o_{Depth}\}$ at each training step (we assume $o_{prop}$ is always present). The latent representation with missing modality is referred to as as $z'_{mult}$. 

We encourage $z'_{mult}$ to be close to our full modality representation $z_{mult}$, so a policy trained on $z_{mult}$ can take $z'_{mult}$ as input. We encourage this by introducing a {\em latent distance loss\/} between them. Our full objective then becomes  $ELBO(o_i, y_i, a_i) + ||z_{mult} - z'_{mult}||^2_2$.

\subsection{Out-Of-Distribution Detection}
\label{sec:ood-detection}
 Similar to \cite{richter2017safe}, we use reconstruction error (L2 distance between input and its reconstruction) to detect input that is out-of-distribution and therefore deemed to come from a corrupted sensor. For some observed modality $o_m \in \{o_{RGB}, o_{Depth}, o_{Force}\}$, we assume that the multimodal model the reconstruction error will be large when the model reconstructs inputs that are OOD from training data. We can threshold the reconstruction error as a method of predicting OOD inputs. For each modality, we choose the thresholds with the best Area Under the Receiver Operating Characteristic Curve (AUROC) performance for detecting corruption in the validation dataset. For the reconstructed F/T data, we threshold the reconstruction error on each of the 3 dimensions and consider the modality an outlier when 2 or more of the dimensions are out-of-distribution. Since we are using a multimodal representation, corruption in one modality may affect reconstruction in another modality leading to the detection of more than one corrupted modality. We handle this ambiguity by selecting the modality with the largest standard deviation away from the mean reconstruction error for that modality in the training set. We found these hyperparameters to be robust in our experiments.

\subsection{Crossmodal Compensation}
\label{sec:crossmodal-compensation}
Following the detection of the corrupted input with OOD detection, we leverage the cross-modal relationships in our model to compensate for the corrupted input. Our approach does not make any assumptions about the nature of the modality corruption. We take the conservative approach and assume that the information of the corrupted modality is too out-of-distribution from our training data and cannot be used. 

We then treat the crossmodal compensation problem as a missing modality problem. Since our representation model is trained with missing modalities, we can simply drop the corrupted modality and use the recalculated multimodal representation $z'_{mult}$ as a proxy for $z_{mult}$, as seen in Fig.~\ref{fig:pull}.

% hile other works in the context of adversarial defense \cite{meng2017magnet} transforms anomalous inputs to cleaned versions, we do not make assumptions about the nature of the corruption and therefore cannot learn a generative model that will work in these instances. Instead, we take the conservative approach and assume that the information in the modality is adversarial and cannot be used. 
% project anomalous inputs to the train distribution, 

For the context of this study, we detect and compensate corrupted sensor readings from one of the three input modalities, namely RGB images, force, and depth.

\subsection{Policy Learning}
\label{sec:policy-control}
% \by{IMO this section can be cut and/or folded into the others -- SAC feels fine to leave as an experimental design detail (which it already is), and the note on what to do when corrupted inputs are detected seems like it belongs in the method overview}
Our final goal is to equip a robot with a policy for performing contact-rich manipulation tasks that leverages our latent multimodal representation.
% We use a state-of-the-art off-policy reinforcement learning algorithm, Soft Actor-Critic~\cite{haarnoja2018soft}, to learn policies for our task.
We use our latent multimodal representation $z_{mult}$ as state input to our policy. When there is a detected corrupted input, the policy takes the crossmodal compensated representation $z'_{mult}$ as state input. 

%% file: 4-expt.tex
\section{Experimental Design}
\begin{figure}
\centering
\includegraphics[width=0.99\linewidth,
trim={0cm 6cm 12.5cm 0cm},clip]{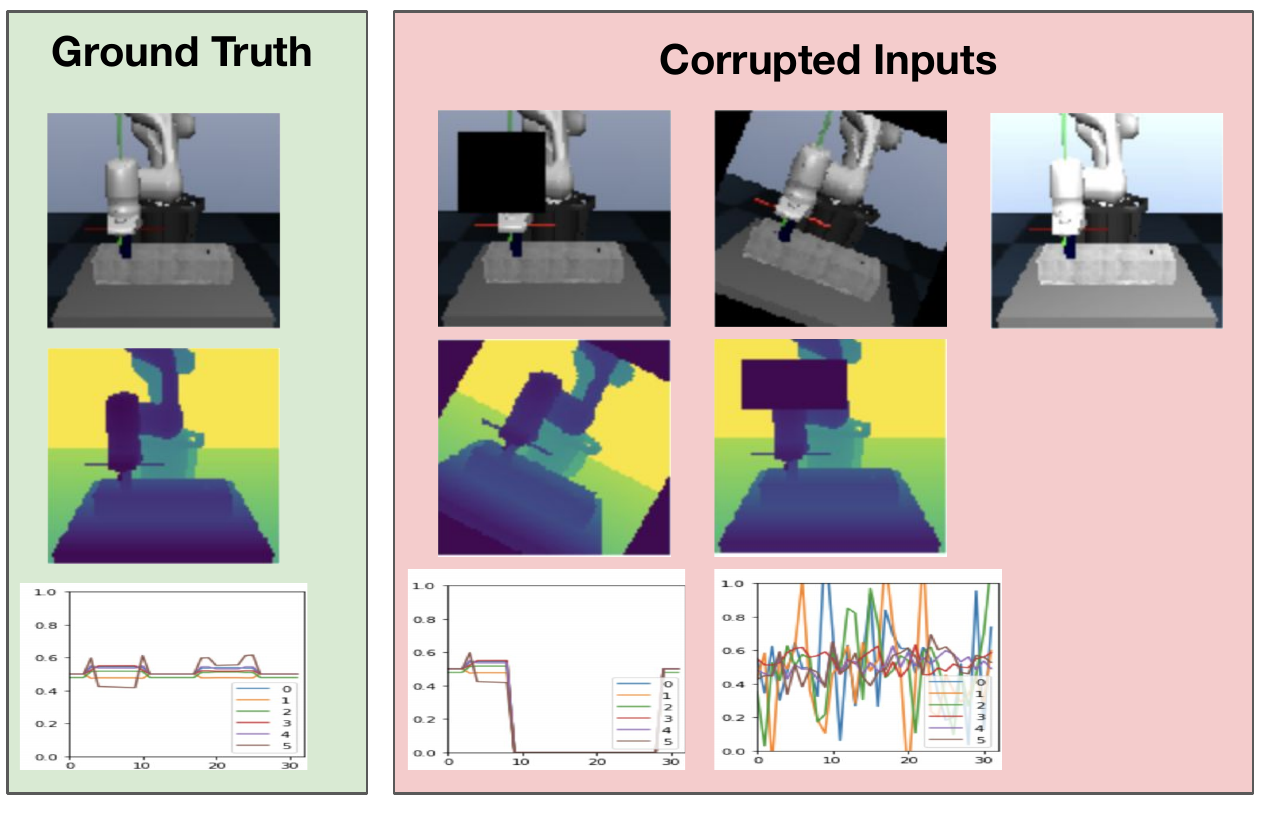}
\caption{Examples of the ground truth data and various kinds of corrupted inputs (from the top and left to right): RGB occlusion, RGB rotations, RGB lighting change, depth rotation, depth box occlusion, blackout force, and force noise. }
\label{fig:ood_examples}
\end{figure}
The primary goal of our experiments is to examine the effectiveness of our pipeline in detecting and compensating for corrupted inputs while retaining useful features for policy learning. Our experiments attempt to answer two main questions: 1) How well can we learn a manipulation task using the learned representation as input? 2) Can a policy trained with CCM handle corrupted sensor input? 

To analyze how our model detects and compensates for corrupted sensor modalities, we additionally design experiments to answer the following questions: 3) Can we use reconstruction error to reliably detect corrupted sensor readings? 4) How close to the latent full-modality representation $z_{mult}$ is the crossmodal compensated representation $z'_{mult}$? 
%\jean{These questions describe more what you will evaluate and even measure. But I would like to see them formulated as hypotheses. For example: Our hypothesis is that reconstruction error can be used for reliably detecting out-of-distribution inputs even for input from multiple modalities. The second points goes more into depth on why policy learning should even worked despite dropping a modality. It kind of backs up the fourth point. And about the third point, will this be more of a reproducing your previous work? Or will it more be an ablation study?}

%\mish{Could we just state the hypothesis in the intro instead? I like how the questions correspond to the experimental metrics, similar to how we did it in the icra paper}

\subsection{Experimental Setup.} 
\label{sec:robot_expt}
\noindent\textbf{Peg Insertion Task} We use a contact-rich peg insertion task as an experimental test-bed for our algorithm.
 We perform our robot experiments in simulation using the RoboSuite \cite{martin2019iros} platform with the Franka Panda robot, a 7-DoF torque-controlled robot. Four sensor modalities are available in simulation, including proprioception, an RGB-D camera, and a force-torque sensor. The proprioceptive input is the end-effector pose as well as linear and angular velocity. RGB images and depth maps are recorded from a fixed camera pointed at the robot. Input images to our model are down-sampled to $128\times 128$. We use a square peg with size (1.4 x 1.4 x 7.5)cm and 1mm clearance in all directions. We use the shaped reward for peg insertion from~\cite{lee2019icra}.

\noindent\textbf{Dataset Collection and Pre-processing.} We apply standard pre-processing techniques for our RGB image and depth inputs by normalizing the data with the min and max values. We clip and normalize the range of F/T sensor data to the 97th percentile and 3rd percentile of the data for each of the 6 dimensions; there is  minimal difference between a robot applying 30N of force and 100N of force for our task. We also weight samples in the dataset to ensure that there are sufficient samples of contact data during training. Finally, for the model to learn pairing, we use unpaired examples where the robot end-effectors are at least 6 centimeters apart from each other to make sure that the inputs are sufficiently different from one other. 

\noindent\textbf{Reinforcement Learning Algorithm and Architecture.} 
After representation training, we freeze the representation model and use our latent representation as state input to our RL policy. We use a state-of-the-art model-free off-policy RL algorithm, Soft Actor-Critic~\cite{haarnoja2018soft}. We use a 2-layer Tanh Gaussian policy that takes as input the 128-dimensional latent representation from our representation model, and produces 3D position displacement $\Delta x$ of the robot end-effector. We adopt the shaped reward from~\cite{lee2019icra} that encourages the peg to be close to the hole and insert. For simplicity, we refer to the policies trained with the CCM representation as CCM policies, and the representation model as simply CCM. We use the same format for our baselines.

\noindent\textbf{Corrupted Sensor Inputs.} 
\label{sssec:expt:ood_pred}
We test our model in detecting and compensating for a variety of unimodal corrupted inputs after policy learning. To accomplish this, we design a wide range of sensor corruptions. For RGB images, we randomly insert black boxes around the robot, change the lighting, and perform in-plane image rotations. For depth inputs, we insert similar random occlusions, and transform the depth image with random in-plane rotations. For F/T inputs, we randomly set forces to 0 (which corresponds to the bottom third percentile of the representation training data) and random Gaussian noise with variances [0.5, 0.25, 0.1]. Examples of different kinds of corrupted inputs are shown in Figure~\ref{fig:ood_examples}.
%\jean{again, I find it setting the F/T values all to zero very weird as it corresponds to actual correct input that should even be in the dataset}

%\mish{ the z dist  for setting F/T values to zero randomly is actually higher than the zdist for 0.25 and 0.1 random noise. I can take it out though and re-evaluate
%}

% \mish{the z_dist fo}
% \mish{the z_dist for setting F/T values to zero randomly is actually higher than the z_dist for 0.25 and 0.1 random noise. I can take it out though and re-evaluate.}

\begin{table*}[]
\centering
\footnotesize
% \resizebox{\linewidth}{!}{%
\begin{tabular}{@{}lccccccc@{}}
\toprule
Model Name & \multicolumn{1}{l}{Normal Input} & \multicolumn{2}{c}{Corrupted Image} & \multicolumn{2}{c}{Corrupted Depth} & \multicolumn{2}{c}{Corrupted Force} \\ \hline &
 & \multicolumn{1}{l}{Comp.} & \begin{tabular}[c]{@{}c@{}}Not  Comp.\end{tabular} & \multicolumn{1}{l}{Comp.} & \begin{tabular}[c]{@{}c@{}}Not  Comp.\end{tabular} & \multicolumn{1}{l}{Comp.} & \begin{tabular}[c]{@{}c@{}}Not  Comp.\end{tabular} \\ \cline{3-8} 
CCM (Our Model) & 96.7\% & \textbf{80.7\%} & 29.3\% & \textbf{82.0\%} & 0.7\% & \textbf{78.0\%} & 81.3\% \\
MFM & \textbf{100.0\%} & 8.7\% & 50.0\% & 0.7\% & 0.7\% & 57.3\% & 18.7\% \\
Recon CCM & 81.3\% & 71.3\% & 69.3\% & 67.3\% & 2.0\% & 69.3\% & 72.0\% \\
% Forward Dynamics CCM & 43.3\% & 40.7\% & 6.0\% & 4.7\% & 2.0\% & 38.7\% & 30.0\% \\
SS CCM & 43.3\% & 40.7\% & 6.0\% & 4.7\% & 2.0\% & 38.7\% & 30.0\% \\
CCM No Dist & 99.3\% & 0.7\% & 28.0\% & 3.3\% & 0.0\% & 30.0\% & 22.0\% \\
CCM No Force & 96.7\% & 78.7\% & 5.3\% & 44.7\% & 14.0\% &   n/a & n/a  \\ 
\bottomrule
\end{tabular}
% }
\caption{The average success rates for our policies. We train 3 policies per model, and evaluate 50 trials per each policy with: normal inputs, corrupted but compensated inputs (Comp.), and corrupted but not compensated inputs (Not Comp.). We see that while MFM had the highest task success rates when given normal inputs, our proposed model, CCM, outperformed all other baselines when compensating for corrupted modality inputs. Not Comp. is given as comparison to see how policies perform when no compensation occurs.}
\label{table:success}
\end{table*}

\noindent\textbf{Implementation Details.} We train the representation model with the Adam optimizer \cite{kingma2014adam} with a learning rate of 0.0001 and $\beta$ values of (0.9, 0.999). The model is trained over 75 epochs with a batch size of 64. For policy learning, we train 3 random seeds per representation model, for 750,000 training steps. Each episode horizon is 200 steps. 

\noindent\textbf{Evaluation Metrics.} We evaluate our algorithm's ability to compensate for corrupted inputs by reporting the success rate of our trained policies when given corrupted depth, image, and force readings. We also report the task success rate of our trained policies when given normal inputs. 

% Our process for compensating for corrupted inputs can be split into two steps: (1) detecting corrupted inputs and (2) throwing out the corrupted modality and compensating for it with our representation model. We look at two metrics to determine the efficacy of these steps.
 
%  We evaluate how well our models can detect corrupted inputs by comparing the Area Under the Receiver Operating Characteristics (AUROC) of the reconstruction errors in predicting whether an input is corrupted. AUROC measures performance across all possible classification thresholds. A model that can perfectly distinguish between in-distribution and OOD examples will have an AUROC of 1.0. An ideal representation should produce a clear reconstruction error boundary that differentiates between clean inputs and corrupted inputs, which is reflected by a higher AUROC. 

% To analyze how each model is compensating for a missing modality, we report the L2 distance between the latent representation given uncorrupted, full modality inputs $z_{mult}$ and the latent representation with missing modality i $z'_{mult, i}$. 

\subsection{Baselines}

We choose to compare CCM with the Multimodal Factorized Model (MFM), another multimodal representation model that deals with missing modalities~\cite{tsai2018learning}. MFM uses Wasserstein Autoencoders (instead of a variational version) to learn a factorized multimodal representation, one for reconstruction objectives and one for self-supervised objectives, and explicitly parameterizes each missing modality model with neural networks. Although MFM was not introduced as a representation model for policy learning and does not perform OOD detection, we include it as part of our baseline model and implement the same corrupted sensor detection algorithm as CCM.  

Our model architecture is similar to the Multimodal Variational Autoencoder (MVAE)~\cite{wu2018multimodal}, but with two key differences: we train with additional self-supervised and latent distance losses. Instead of considering MVAE as a baseline, we evaluate how the different components of CCM contribute to its success with an ablation study. 

\subsection{Ablation Study}

A full CCM model learns a representation with self-supervised, reconstruction, and latent distance objectives. We propose the following ablation baselines: 

% Because force has very different characteristics from depth and image data, we also study the role force plays in crossmodal compensation by including a model trained without force during representation and policy learning. 

\begin{enumerate}
    \item \textbf{Recon CCM:} CCM trained with reconstruction and latent distance objective. 
    \item \textbf{SS CCM:} CCM trained with self-supervised and latent distance objectives. We use Recon CCM for corrupted sensor detection.
    % Because SS CCM cannot perform reconstruction, this baseline uses a separately learned Recon CCM for out-of-distribution detection. 
    \item \textbf{CCM No Dist:} CCM trained with self-supervised and reconstruction objectives only. 
    \item \textbf{CCM No Force:} CCM trained with zeroed out forces during representation and policy learning. 
    
    % \mish{make sense to have CCM No depth and CCM no image?}
    % \msal{Idk, would be cool to know how impt depth or rgb are relative to each other, but doesn't feel necessary to the point of the paper}
    
\end{enumerate}

\section{Experimental Results and Discussion}

\subsection{Policy Learning for Contact-rich Tasks}
\begin{figure}

\begin{subfigure}[b]{0.49\textwidth}
\centering
\includegraphics[width=\linewidth,trim={0.0cm 0.0cm 0.35cm 1.25cm},clip]{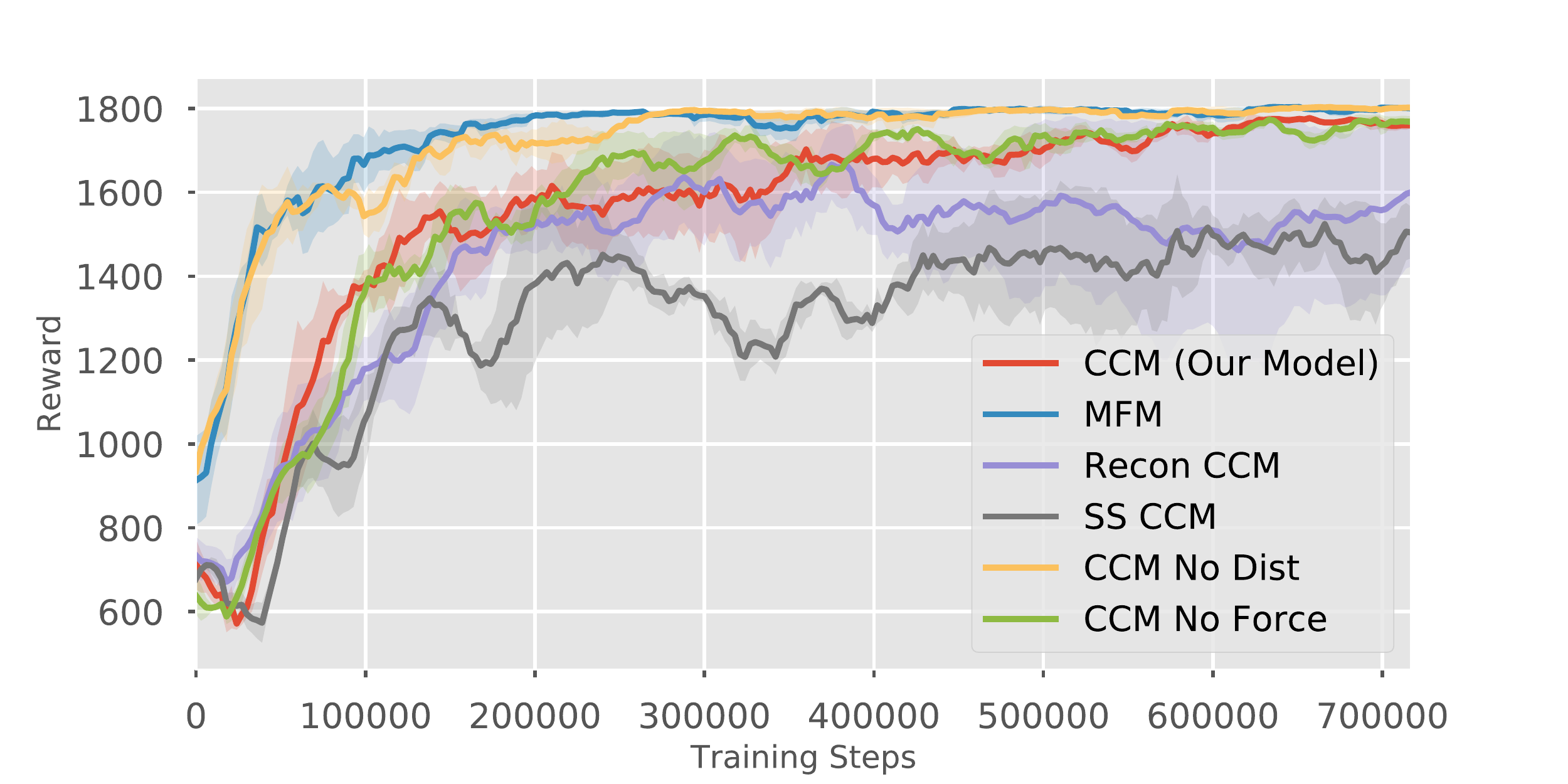}
\caption{Policy training curve}
\label{fig:rl}
\end{subfigure}
% trim=left bottom right top, clip
\begin{subfigure}[b]{0.49\textwidth}
    \centering
\includegraphics[width=\linewidth,
trim={0.20cm 0.0cm 0.35cm 1.25cm},
clip]{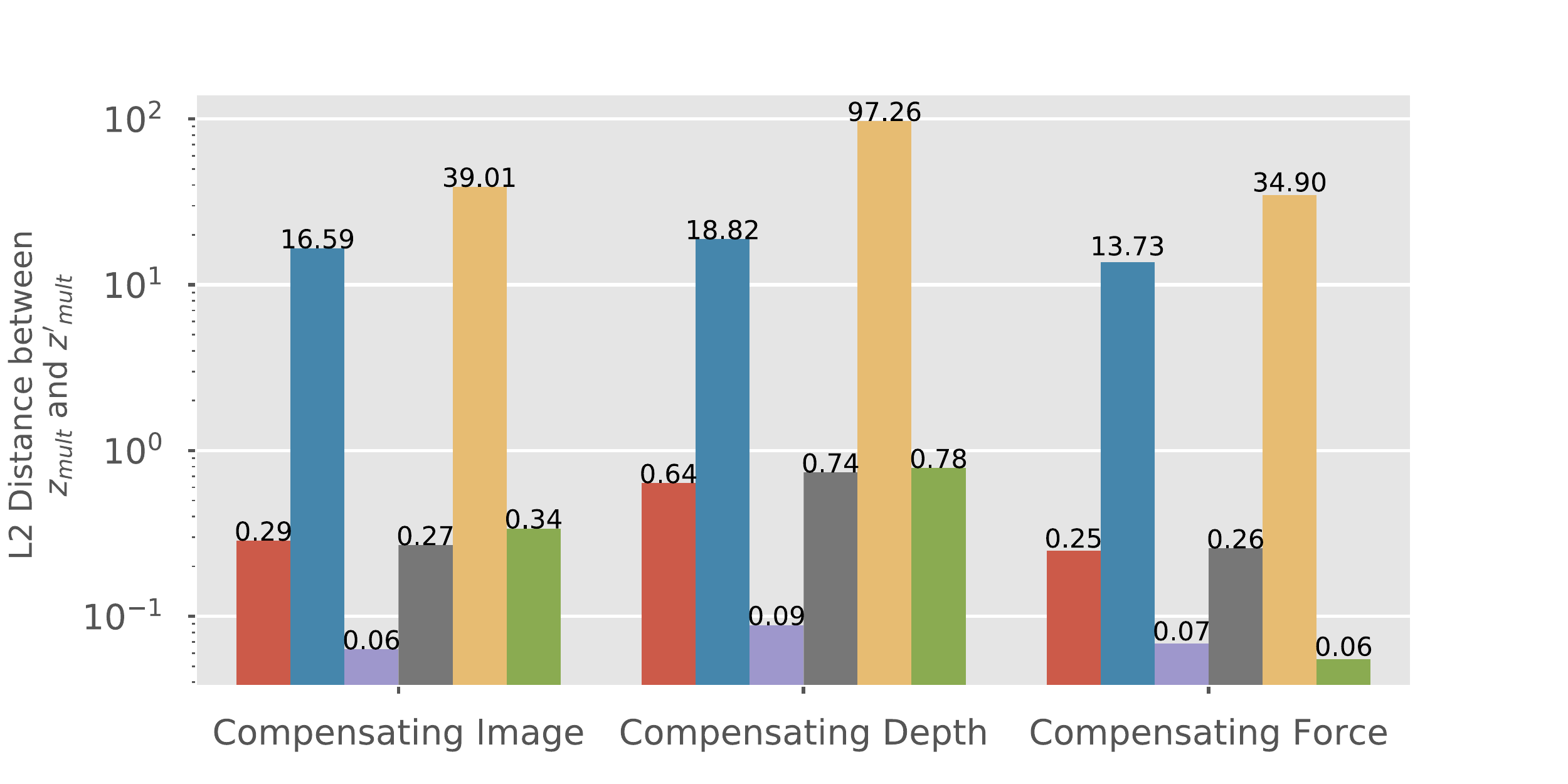}
\caption{Latent L2 Distance}
% \caption{L2 distance in the latent space for compensated modalities }
\label{fig:distance}
\end{subfigure}
\caption{(a) Training curves for reinforcement learning on the peg insertion task, 3 random seeds per model (b) We show in log-scale the latent L2 distance between full modality $z_{mult}$ and compensated modality $z'_{mult}$, for different compensated modalities. CCM No Dist and MFM had relatively high latent L2 distances, providing an explanation why the two models struggled with compensation.}
% \yz{(a) took a lot of space but had not a lot of information, use log-scale, shorter figure, and bigger font sizes, or replace with a table} \jean{personally I think it is good to have take-away messages in the captions. You can also make the font size of the captions smaller}
\end{figure}

We report the learning curves of our policies for the peg insertion task in Fig.~\ref{fig:rl}. To evaluate the success rate of peg insertion given normal inputs, we evaluate every learned policy 50 times, and report for each model the average policy success rate in Table~\ref{table:success} (see Normal Input column). 

The MFM policies are able to learn the task faster, and had a 100\% success rate given normal inputs, compared to CCM policies' 96.7\% success rate. By factorizing the self-supervised and reconstruction models and learning a new neural network model for each missing modality, MFM has a more complex model architecture than CCM and more parameters (7.7 million compared to 5.6 million), which might explain its success in learning the task. 

For our ablation study, we see that CCM No Dist policies also learned as fast as the MFM policies, and achieved a 99.3\% success rate. In comparison, SS CCM and Recon CCM policies had high variances among its learned policies, and had low average task success rates of 43.3\% and 81.3\% respectively. CCM policies' better performance suggests that learning both self-supervised and reconstruction objectives helps to learn a more successful representation for policy learning. On the other hand, the success of CM No Dist policies suggests that learning the latent distance objective negatively affects the representation's efficacy as state representation for controls. One possible explanation is that when we drop a modality during training, the distance objective encourages the latent representation to only learn features that can be crossmodally compensated by the other modalities, which serves as a kind of regularization. However, we only observe a small 2.6\% difference in task success between CCM and CCM No Dist. 

\subsection{Compensation of Corrupted Inputs}
We evaluate each learned policy 50 times for corrupted force, image, and depth inputs. For each type of corrupted modality, we randomly choose a type of sensor corruption associated with the modality (described in Sec.~\ref{sec:robot_expt} and Fig.~\ref{fig:ood_examples}) at every step of the policy rollout. The task success rates can be found in Table~\ref{table:success}. We also report the average success rate when sensor inputs are corrupted but not compensated for.

The MFM policies struggled with compensating corrupted inputs, especially for image and depth data, in which the success rates dropped from 100\% to 8.7\% and 0.7\% respectively. 

CCM No Dist policy success rates also dropped drastically when given corrupted inputs. This is expected, because CCM No Dist is not trained to map the compensated $z'_{mult}$ to the full modality $z_{mult}$. Thus, the policies receive very different state feedback at test time with corrupted input compared to training with normal input. Both SS CCM and Recon CCM policies performed better than CCM No Dist when given corrupted inputs. However, the performances of SS CCM and Recon CCM in this experiment are upper-bounded by how well the policies trained with normal inputs.

Overall, CCM policies outperformed all others with corrupted sensor inputs. To better understand why CCM representations performed better than the others in crossmodal compensation, we analyze how well each model detects and compensates for corrupted inputs below. 
\subsection{Corrupted Reading Detection}

To study detection of corrupted sensor data, we use corruptions shown in Fig.~\ref{fig:ood_examples} and evaluate the AUROC for detecting them in the representation learning test set. As seen in Table~\ref{table:auc}, all models were able to detect corrupted depth data with an AUROC score of 1.0. Both MFM and CCM No Dist have lower AUROC scores for detecting corrupted Image and Force data than other models, but not by much. 

Because CCM No Dist and MFM both had high AUROC for corrupted depth data detection, but failed to compensate for corrupted depth data during policy rollout, this suggests that CCM No Dist and MFM struggled with compensation rather than detection of corrupted inputs. 

% As a note, SS CCM could not perform modality corruption detection as the model does not learn reconstruction of input modalities, but this baseline uses the Recon CCM model to perform corrupted modality detection.

\subsection{Latent Representation Distance}
We show the L2 distance between the full modality latent variable $z_{mult}$ and the compensated latent variable $z'_{mult}$ in Fig.~\ref{fig:distance} as evaluated on the test set for representation learning. While a small L2 distance does not guarantee successful crossmodal compensation, it is a proxy measure for how well a model can compensate for corrupted modalities during policy rollout. 

Notably, MFM and CCM No Dist have much higher latent distances than others. This explains their poor performance in compensating for corrupted inputs. Recon CCM had the lowest latent distances, explaining why Recon CCM policies had lower performance drop with corrupted inputs than CCM policies (on average, a 12\% vs. a 16.5\% drop). However, CCM policies had better normal input policy performance. In other words, the representations learned with CCM are able to balance good policy performance with normal inputs as well as crossmodal compensation with corrupted inputs. 

\subsection{Redundant Information among Modalities}

Past works have demonstrated the ability to predict haptic information from vision and vice versa~\cite{li2019connecting, takahashi2019deep}, indicating that force and visual data share redundant information. In our results, we observe that CCM policies performed 3.3\% better when using corrupted force input than when compensating for force, and Recon CCM performed 2.7\% better when using corrupted force input than when compensating for it. This suggests the policies might be ignoring the force input. 

Although these results show that the benefit of compensating for corrupted force information is limited when vision data is available, the results from our CCM No Force baseline show that force information helps compensate for corrupted depth and image inputs. CCM No Force has lower task success rates compared to CCM: 44.7\% compared to 82\% when compensating for corrupted depth, and 78.7\% compared to 80.7\% when compensating for corrupted images. It also results in higher latent distances when either visual input is missing.

\begin{table}[]
\centering

\begin{tabular}{@{}llll@{}}
\toprule
Models             & Depth         & Image         & Force         \\\midrule
CCM (Our Model)    & \textbf{1.00} & \textbf{0.98} & 0.97          \\
MFM                & \textbf{1.00} & 0.97          & 0.89          \\
Recon CCM & \textbf{1.00} & \textbf{0.98} & \textbf{0.98} \\
CCM No Dist & \textbf{1.00} & 0.95          & 0.92          \\
CCM No Force       & \textbf{1.00} & 0.97          &   n/a           \\    \bottomrule
\end{tabular}
\caption{AUROC for classifying if multimodal inputs had corrupted depth, image, or force data in our test data. We used a threshold reconstruction loss (comparing input data and reconstructed data) for each modality to perform corruption detection.}
% \yz{not sure what you try to show here, but a table with all results closer to perfect looks fishy in general.} \jean{Can you be more explicit in the caption? What is the take-away?}
% \msal{I don't think it looks fishy for this problem. Agree with adding more context though.}
\label{table:auc}
\end{table}

%% file: 5-disc.tex
\section{Conclusions}

We introduced CCM, a self-supervised method for learning representations that crossmodally compensatse for corrupted inputs. By leveraging reconstruction losses, CCM can detect a variety of corrupted sensor inputs. Following detection, CCM rejects and discards the corrupted modality and use the remaining modalities to approximate the joint multimodal representation through crossmodal compensation. We showed that the policies learned with the CCM's representation is able perform a peg insertion task even when sensor inputs are corrupted. We compare CCM with other multimodal representation learning baselines, and perform a thorough analysis of how our model performs in detecting corrupted sensor inputs and compensating for them. We find that our novel model outperforms all others for task completion with corrupted sensors.